\documentclass[fleqn,10pt,twocolumn]{wlscirep}
\usepackage{subcaption}
\usepackage{threeparttable}
\usepackage{graphicx}
\usepackage{booktabs}
\usepackage{fontspec}
\usepackage{libertine}
\usepackage{amsmath, amssymb, amsthm, mathrsfs,bm}
\usepackage{booktabs}
\usepackage{url}
\usepackage{graphicx}
\usepackage{subcaption}
\usepackage{enumitem}
\usepackage{algorithm}
\usepackage{algpseudocode}
\usepackage{comment}
\usepackage{multirow}
\usepackage{xcolor}
\title{Artificial Intelligence in Breast Cancer Care: Transforming Preoperative Planning and Patient Education with 3D Reconstruction}

\author[1]{Mustafa Khanbhai}
\author[1, 4]{Giulia Di Nardo}
\author[2]{Jun Ma}
\author[3]{Vivienne Freitas}
\author[1]{Caterina Masino}
\author[4]{Ali Dolatabadi}
\author[1]{Zhaoxun ``Lorenz'' Liu}
\author[3]{Wey Leong}
\author[1]{Wagner H. Souza}
\author[1]{Amin Madani}
\affil[1]{Surgical Artificial Intelligence Research Academy, University Health Network, Toronto, ON, Canada}
\affil[2]{Vector Institute, Toronto, ON, Canada}
\affil[3]{Princess Margaret Cancer Centre, University Health Network, Toronto, ON, Canada}
\affil[4]{Department of Mechanical \& Industrial Engineering, University of Toronto, Toronto, ON, Canada}

\begin{abstract}

\textit{Background}

Effective preoperative planning and decision-making require accurate and generalizable algorithms capable of segmenting complex anatomical structures and detecting pathological features across various datasets. However, traditional machine learning models often struggle with generalizing across diverse patient data, leading to limitations in accuracy and usability. This study presents a novel machine learning methodology designed to improve the generalization of algorithms for 3D anatomical reconstruction, demonstrating the potential of these models beyond breast cancer to a wide range of medical imaging applications. 

\vspace{5pt}
\textit{Methods}

A total of 120 retrospective breast MRIs from January 2018 to June 2023 were processed in three phases: (1) anonymization, pre-processing, and manual segmentation of T1-weighted (T1W) and dynamic contrast-enhanced (DCE) sequences; (2) co-registration of T1W and DCE images, followed by segmentation of whole breast, fibroglandular tissue and tumour; (3) visualization of merged 3D models using ITK-SNAP. A human-in-the-loop approach refined segmentations for both breast anatomy and tumor lesions using U-Mamba, a machine learning (ML) model designed to generalize across various imaging scenarios. The Dice similarity coefficient (DSC) was used to assess overlap between automated segmentation and ground truth, with values ranging from 0 (no overlap) to 1 (perfect overlap), to evaluate the algorithm's ability to generalize across datasets. Clinical relevance of the final 3D models was evaluated through interviews with clinicians and patients. 

\vspace{5pt}
\textit{Results}

A total of 120 MRIs were segmented and analyzed, with the U-Mamba model showing strong performance across the datasets. For T1-weighted (T1W) images, the DSC for anatomical structures such as whole organs and fibroglandular tissue was 0.97 (+0.013) and 0.96 (+0.024), respectively. For pathological regions (e.g., tumors), the DSC was 0.82 (+0.12). The model was successful in generating accurate 3D reconstructions that allowed clinicians to visualize complex anatomical features across different cases. Interviews with clinicians indicated that the use of these models improved planning, intraoperative navigation, and overall decision support. Furthermore, the integration of 3D visualization into the patient education process was cited as enhancing communication, enabling better patient understanding and decision-making. 

\vspace{5pt}
\textit{Conclusion}

This study demonstrates the successful application of a novel, human-in-the-loop machine learning approach for generalizing algorithms in the context of 3D reconstruction and anatomical segmentation. By improving the algorithm's ability to generalize across different patient datasets, this approach holds promise for a wide range of medical applications, beyond just breast cancer. The generated 3D models offer enhanced visualization for clinicians, improved preoperative planning, and more effective patient education, facilitating shared decision-making and empowering patients to make informed choices about their care.

\end{abstract}
\begin{document}

\flushbottom
\maketitle

\thispagestyle{empty}

\section{Background}\label{Background}

Breast conserving surgery (BCS), though widely used for early-stage breast cancer, presents planning challenges due to variability in breast anatomy, tumor size, and location.\cite{paulinelli2020oncoplastic} Surgical decisions often rely on subjective clinical judgment, lacking objective tools or predictive models.\cite{qu2022preoperative}  Current surgical planning in  BCS primarily relies on the surgeon’s ability to interpret and mentally reconstruct two-dimension slices of computed tomography/magnetic resonance (CT/MR) images into three-dimension (3D) while planning surgical resection. This is mentally strenuous and increases cognitive load, as it relies on CT/MR interpretation, anatomical and surgical knowledge, experience, and spatial sense.\cite{hattab2021investigating, yeo2018utility} Patients frequently face the decision unprepared, \cite{fagerlin2006informed} with limited understanding \cite{fernandes2011post, schulman2020patient} of long-term outcomes and few resources to visualize surgical options,\cite{davies2024understanding} contributing to decision regret and dissatisfaction\cite{killoran2006surgical, advani2019local}.  

Recent advancements in machine learning (ML) have enabled new approaches to process and interpret complex medical imaging data \cite{brown2023breaking}. Visual tools such as patient-specific 3D breast models \cite{bessa20203d, duraes2019surgery} have shown potential to improve preoperative planning. This could provide more objective and time-efficient methods assisting in surgeons’ pre-operative planning, reduced mental projection of two-dimension (2D) images onto a 3D surgical scenario, increased accuracy of tumour location, and shared decision-making through patient education \cite{duraes2019surgery, faermann2014tumor}. ML algorithms could potentially streamline the process of autonomous breast imaging segmentation (including normal anatomy and tumour) to detect and localize breast lesions on radiological examinations, of which, magnetic resonance imaging (MRI) stands as a pivotal imaging modality to detect breast cancer.\cite{mann2019contrast}

This study proposes a novel ML-based methodology for generating high-fidelity 3D breast reconstructions from breast MRI data using a human-in-the-loop segmentation pipeline for segmentation of breast anatomy and tumor structures. The approach is designed to generalize across diverse imaging inputs and patient profiles. In addition to evaluating the technical performance of the model, we assess its potential for enhancing clinical workflows, surgical planning, and patient education. 

\section{Methods}\label{Methods}

Lesion identification in breast MRI has shown promise with the rapid advancement of deep learning, where convolutional neural networks (CNNs) have demonstrated high performance to automate the segmentation of various anatomical regions within breast MRI \cite{doran2017breast, zhang2022automatic, lew2024publicly, guo2022breast, ham2023improvement}. The high level of agreement between human- and machine-generated segmentation maps demonstrates the potential of CNNs. A recent study \cite{muller2023fibroglandular} demonstrated that transformer-based networks have further improved the quality of breast segmentation in breast MRI compared to CNNs like nnUNet. While Transformers have improved the ability of long-range dependencies, they are generally very computationally expensive. In this study, we use U-Mamba \cite{ma2021umamba} to address the challenges in modelling long-range dependencies. Extensive experiments demonstrate U-Mamba achieves superior performance than both CNN- and Transformer-based networks on four diverse tasks.\cite{ma2021umamba}

This 24-month study received ethical approval from the University Health Network (UHN) Research Ethics Board, 23-5533.0. The study was conducted in three phases as outlined below. Retrospective breast MRIs from UHN were retrieved from 1st January 2018 to 15th June 2023. Breast MRI was performed on either a 1.5-T (Siemens Healthiners) or a 3.0-T (Siemens Healthiners) scanner using a 16-channel breast coil. 

\textit{\textbf{Phase 1}} 

Breast MRIs of patients that fulfilled the inclusion criteria (Appendix 1) were anonymized and downloaded as Digital Imaging and Communications in Medicine (DICOM) files. The differential diagnosis for lesions detected at breast MRI is based on the morphology of the lesion, perfusion kinetics, and inherent signal characteristics on T1 and T2-weighted sequences following the Breast Imaging Reporting and Data System (BI-RADS) lexicon.\cite{dorsi2013acr} Anonymized MRIs underwent pre-processing, noise reduction, and contrast enhancement prior to manual segmentation. Two MRI sequences from the same patient were downloaded separately: 

\begin{enumerate}
    \item T1-Weighted (T1W) sequence without fat saturation where breast adipose tissue appears bright and while fibro-glandular tissue appears dark.
    \item Dynamic Contrast-Enhanced (DCE) sequence which demonstrates location and size of lesion.
\end{enumerate}

\textit{\textbf{Phase 2}}

Co-registration of both T1W and DCE was performed prior to segmentation. This process aligns the images in both sequences so that corresponding pixels may be integrated. Segmentation of the two sequences is done separately (T1W for breast anatomy and DCE for tumour identification) before both segmentations are combined to produce a complete 3D breast segmentation model.  

We used manual segmentations from publicly available internal and external breast MRI datasets from Duke University \cite{saha2021dynamic} to automate our segmentations. Pre-operative dynamic contrast enhanced (DCE)-MRI: downloaded from PACS systems and de-identified for The Cancer Imaging Archive (TCIA) release. These include axial breast MRI images acquired by 1.5T or 3T scanners in the prone positions. Following MRI sequences shared in DICOM format: a non-fat saturated T1-weighted sequence, a fat-saturated gradient echo T1-weighted pre-contrast sequence, and mostly three to four post-contrast sequences. We used extracted features of breast volume and fibroglandular tissue segmented from the T1-non-fat saturated first post-contrast sequence processed with N4-ITK. Further details of annotation can be found here \cite{saha2021dynamic}. This dataset was implemented using U-Mamba to segment the whole breast outline (distinguished from the background and chest wall) and fibro-glandular tissue (distinguished from the breast adipose tissue) on our 210 T1W non-fat saturated MR images. We developed a human-in-the-loop approach for segmentation. This approach trains an accurate prediction model by integrating human knowledge and experience \cite{wu2022survey}. Segmentation was edited using ITK-SNAP Version 4.0.1. Segment editor was used to edit segmentations on each axial slice for fibroglandular tissue and whole breast outline manually by a breast clinician (MK) with 5 years of experience. The output was used to re-train the U-mamba model which was split into training and test of 75:25. 

\begin{figure*}[t]
\centering
\includegraphics[width=\textwidth]{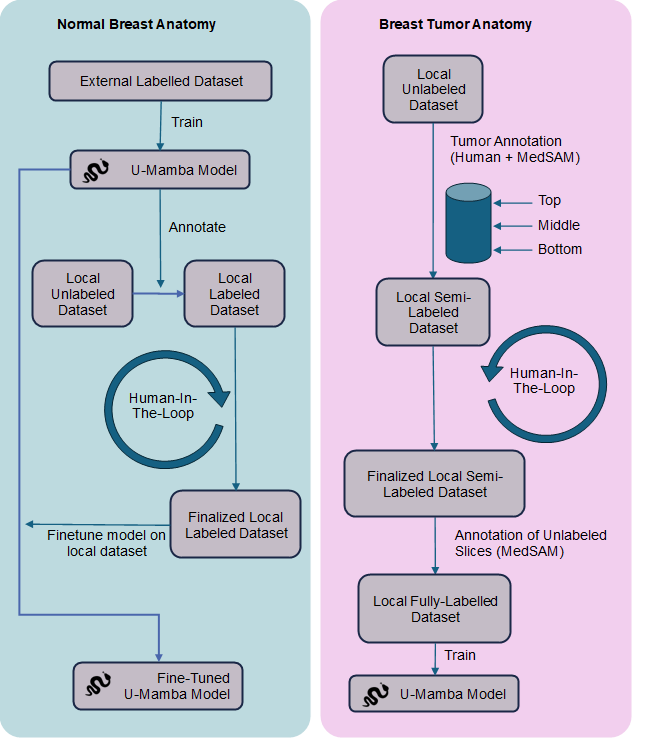}
\caption{Workflow for breast segmentation and machine learning (ML) model development. The process begins with two MRI sequences obtained from the same patient. The images undergo preprocessing steps, including alignment and normalization, followed by segmentation of the breast tissue. A machine learning model is then developed using the segmented data, and iterative refinement is performed to optimize its accuracy and performance.}
\label{fig:workflow}
\end{figure*}

For DCE, we used a semi-automated segmentation method. Three-slice (top, middle, and bottom slices) of each breast lesion was manually annotated using ITK-SNAP Version 4.0.1. Three distinct axial images corresponding to the top, middle (widest diameter) and bottom of the tumour were marked for consistency. MedSAM was used to generate the complete 3D masks. MedSAM is a foundation model designed to bridge this gap by enabling universal medical image segmentation.\cite{ma2024segment} Human-in-the-loop strategy was used to manually revise the masks (directly overwriting the current masks during revision) by the same breast clinician as above. The revised masks were used to train a U-Mamba model for automatic tumour lesion segmentation which was split into training and test of 75:25. The whole process is summarised in Figure \ref{fig:workflow}. Performance metrics (dice similarity coefficient and normalised surface distance) was used to calculate optimisation. The Dice score quantifies the overlap between the automated segmentation and the ground truth, with values ranging from 0 to 1, where 1 represents a perfect overlap. Both final U-Mamba models for T1W and DCE were combined to create a 3D segmentation of the whole breast, fibroglandular tissue, and tumour.  

\textit{\textbf{Phase 3}}

The 3D visualization of the breast was generated using the combined U-mamba 2D segmentation results on ITK-SNAP Version 4.0.1 using the “3D Toolbox”. The 3D image contrast was adjusted using the windowing dialogue for adjusting the mapping between intensities in the 3D image and intensities of the displayed slices. Finally, the SNAP User Interface was used to adjust the appearance and improve the visibility of breast tumour, fibroglandular and adipose tissue.  

To assess the clinical relevance of these 3D breast models, two semi-structured interviews were conducted. Four anonymized 3D breast models were provided to both clinicians and patients. The first survey (Appendix 1) involved breast surgical oncologists, who were asked to identify facilitators and barriers to the clinical implementation and workflow integration of the 3D models. For clinician interviews, a member of the research team approached breast surgical oncologists at UHN with an e-mail invite to participate. Consent was taken prior to one-on-one interviews. The patient survey (Appendix 2) targeted patients who had completed their treatment and evaluated whether the 3D model helped them better understand the tumour’s location and size in relation to the breast and the proposed surgery. Patients were approached in the breast clinic by a member of the research team and a patient information leaflet was provided. If they agreed to participate, their consent was taken prior to conducting one-to-one interviews at the end of their clinic appointment. No patient demographic or clinicopathologic details were collected. 

Qualitative data was analyzed using the Framework analysis method.\cite{pope2000qualitative} The five stages to framework analysis were: 

\begin{enumerate}
    \item Familiarization – The process began by thoroughly reading the completed questionnaires to gain an overall understanding of participant responses. 
    \item Thematic Framework Development – Key themes were identified from the responses, and an initial thematic framework was created based on these recurring patterns. 
    \item Indexing – Participant quotes relevant to the identified themes and subthemes were extracted from the questionnaires and organized into a document. 
    \item Charting – The initial framework was reviewed and refined iteratively. This led to the development of a final thematic framework with main themes. 
    \item Mapping and Interpretation – A summary of the main descriptive comments was created for each theme that provided deeper insights into the participants’ perspectives. 
\end{enumerate}

Thematic saturation was reached when repetitive themes were identified and no new information emerged, and no further participants were recruited. Approval for the study was obtained from the Quality Improvement Review Committee (QIRC) at UHN under protocol 24-0828. In collaboration with the Techna Technology Development Team at UHN, these qualitative insights were utilized to determine the optimal alignment of the 3D model with existing workflows.

\begin{table*}[t]
\centering
\caption{Patient characteristics included in the study, with corresponding breast MRI data utilised for segmentation purposes.}
\label{tab-1}
\begin{tabular}{lc}
\toprule
\textbf{Characteristic} & \textbf{n} \\
\midrule
Mean age (±SD) & 53 (38 ± 75) \\
\midrule
\textbf{Molecular subtype} & \\
\quad Luminal A & 69 \\
\quad Luminal B & 25 \\
\quad Her2 positive & 14 \\
\quad TNBC & 12 \\
\midrule
\textbf{BIRADS Category} & \\
\quad 6 & 120 \\
\midrule
\textbf{Tumour size (cm)} & \\
\quad <2cm & 74 \\
\quad 2cm -- 5cm & 39 \\
\quad >5cm & 7 \\
\bottomrule
\end{tabular}
\begin{tablenotes}
\small
\item HER2: Human Epidermal growth factor receptor 2, TNBC: Triple negative breast cancer, BIRADS: Breast Imaging Reporting and Data System
\end{tablenotes}
\end{table*}

\begin{table*}[t]
\centering
\caption{The U-Mamba model applied to T1W Breast MRI samples, incorporating both the whole breast outline and fibroglandular tissue, achieved strong performance metrics, as indicated by high Dice Similarity Coefficient (DSC) and Normalized Surface Distance (NSD) values.}
\label{tab-2}
\begin{tabular}{llll}
\toprule
\multicolumn{2}{c}{\textbf{Mean Dice similarity coefficient (±SD)}} & \multicolumn{2}{c}{\textbf{Mean Normalized surface distance (±SD)}} \\
\cmidrule(lr){1-2} \cmidrule(lr){3-4}
\textbf{Whole breast outline} & \textbf{Fibroglandular tissue} & \textbf{Whole breast outline} & \textbf{Fibroglandular tissue} \\
\midrule
0.97 (±0.013) & 0.96 (±0.024) & 0.94 (±0.018) & 0.98 (±0.003) \\
\bottomrule
\end{tabular}
\end{table*}

\begin{table*}[t]
\centering
\caption{The U-Mamba model applied to DCE Breast MRI samples, incorporating breast lesions, achieved strong performance metrics, as indicated by high Dice Similarity Coefficient (DSC) and Normalized Surface Distance (NSD) values.}
\label{tab-3}
\begin{tabular}{p{3cm}cc}
\toprule
& \textbf{Mean Dice similarity coefficient (±SD)} & \textbf{Mean Normalized surface distance (±SD)} \\
\midrule
\textbf{Breast Lesion Segmentation} & 0.82 (±0.12) & 0.92 (±0.11) \\
\bottomrule
\end{tabular}
\end{table*}

\section{Results}\label{Results}

According to the selection criteria, a total of 120 Breast MRIs were segmented based on the anatomical areas in the T1W and DCE sequence. Patient characteristics are tabulated in Table \ref{tab-1}. 

\textit{\textbf{T1W}}

\begin{figure*}[t]
\centering
\begin{subfigure}[b]{\textwidth}
    \centering
    \begin{subfigure}[b]{0.24\textwidth}
        \centering
        \includegraphics[width=\textwidth]{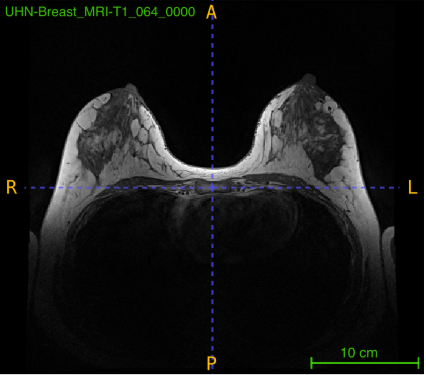}
    \end{subfigure}
    \hfill
    \begin{subfigure}[b]{0.24\textwidth}
        \centering
        \includegraphics[width=\textwidth]{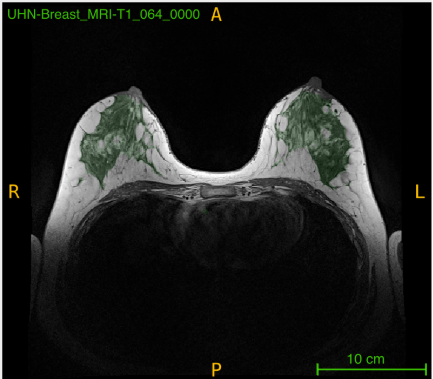}
    \end{subfigure}
    \hfill
    \begin{subfigure}[b]{0.24\textwidth}
        \centering
        \includegraphics[width=\textwidth]{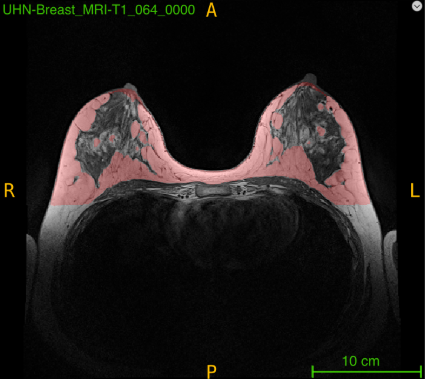}
    \end{subfigure}
    \hfill
    \begin{subfigure}[b]{0.24\textwidth}
        \centering
        \includegraphics[width=\textwidth]{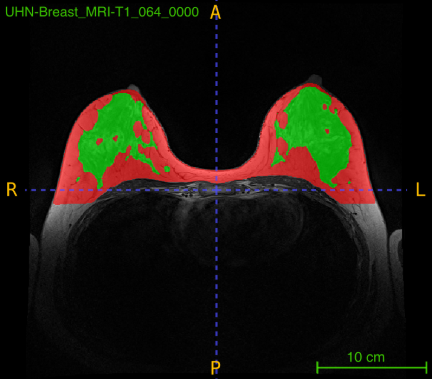}
    \end{subfigure}
    \caption{Patient A}
    \label{fig:2a}
\end{subfigure}

\vspace{0.5em}

\begin{subfigure}[b]{\textwidth}
    \centering
    \begin{subfigure}[b]{0.24\textwidth}
        \centering
        \includegraphics[width=\textwidth]{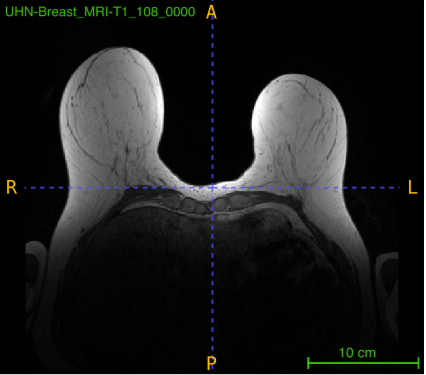}
    \end{subfigure}
    \hfill
    \begin{subfigure}[b]{0.24\textwidth}
        \centering
        \includegraphics[width=\textwidth]{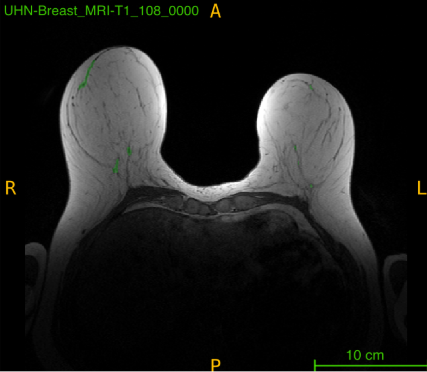}
    \end{subfigure}
    \hfill
    \begin{subfigure}[b]{0.24\textwidth}
        \centering
        \includegraphics[width=\textwidth]{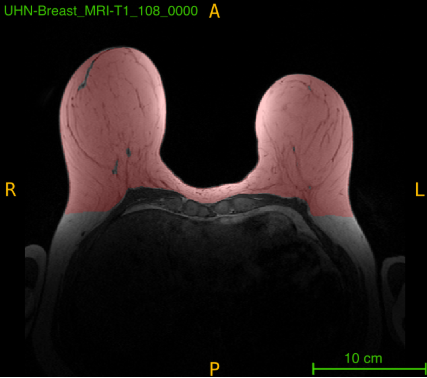}
    \end{subfigure}
    \hfill
    \begin{subfigure}[b]{0.24\textwidth}
        \centering
        \includegraphics[width=\textwidth]{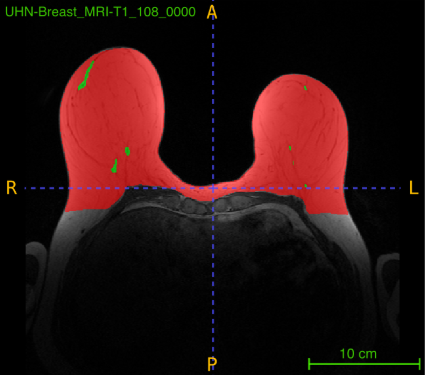}
    \end{subfigure}
    \caption{Patient B}
    \label{fig:2b}
\end{subfigure}

\caption{T1W MRI samples from two anonymized patients illustrating distinct variations in the quantity of fibroglandular tissue. The segmentation results, generated by the U-Mamba model, delineate the fibroglandular tissue regions (green) and whole breast (red), effectively capturing these differences between the two cases.}
\label{fig-2}
\end{figure*}

The output of the final U-Mamba segmentation for T1W with whole breast outline and fibroglandular tissue showed strong performance metrics, as indicated in Table \ref{tab-2}. Mean DSC (+Standard Deviation [SD]) for whole breast outline and fibroglandular tissue was 0.97 (+0.013) and 0.96 (+0.024), respectively. MRI samples from two anonymized patients highlighted noticeable variations in the quantity of fibroglandular tissue present (Figure \ref{fig-2}). These differences were clearly reflected in the segmentation results, where the U-Mamba model delineated the fibroglandular tissue regions for both cases. 

\textit{\textbf{DCE}}

\begin{figure*}[t]
\centering
\begin{subfigure}[b]{\textwidth}
    \centering
    \begin{subfigure}[b]{0.3\textwidth}
        \centering
        \includegraphics[width=\textwidth]{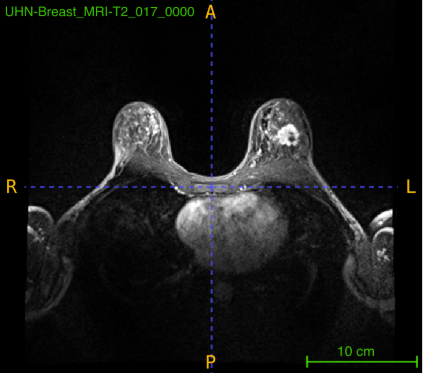}
        \caption*{i}
    \end{subfigure}
    \hfill
    \begin{subfigure}[b]{0.3\textwidth}
        \centering
        \includegraphics[width=\textwidth]{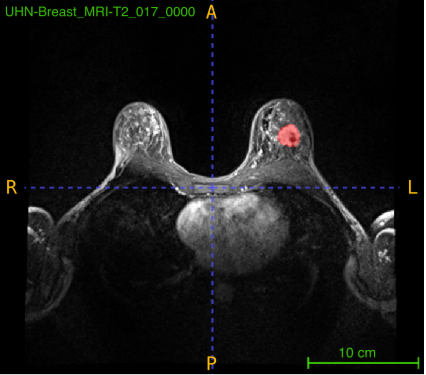}
        \caption*{ii}
    \end{subfigure}
    
    \vspace{0.3em}
    
    \begin{subfigure}[b]{\textwidth}
        \centering
        \includegraphics[width=0.8\textwidth]{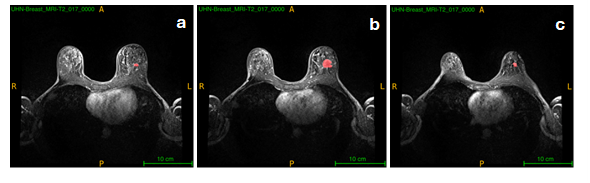}
    \end{subfigure}
    
    \caption{Patient A}
    \label{fig:3a}
\end{subfigure}

\vspace{0.8em}

\begin{subfigure}[b]{\textwidth}
    \centering
    \begin{subfigure}[b]{0.3\textwidth}
        \centering
        \includegraphics[width=\textwidth]{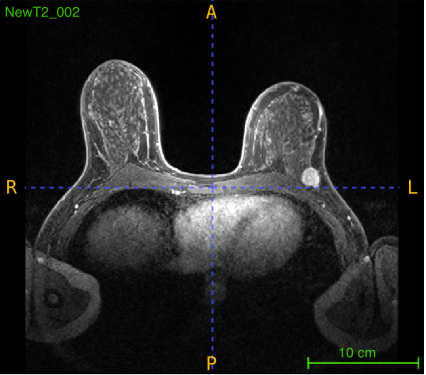}
        \caption*{i}
    \end{subfigure}
    \hfill
    \begin{subfigure}[b]{0.3\textwidth}
        \centering
        \includegraphics[width=\textwidth]{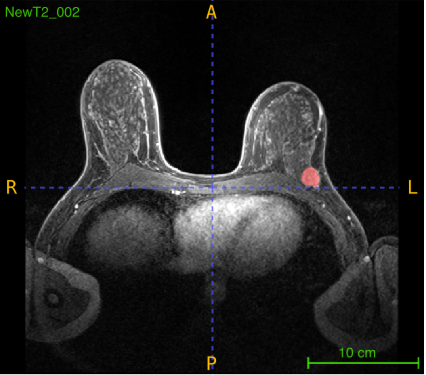}
        \caption*{ii}
    \end{subfigure}
    
    \vspace{0.3em}
    
    \begin{subfigure}[b]{\textwidth}
        \centering
        \includegraphics[width=0.8\textwidth]{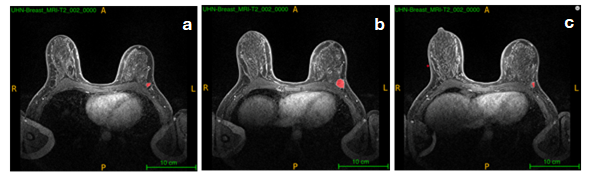}
    \end{subfigure}
    
    \caption{Patient B}
    \label{fig:3b}
\end{subfigure}

\caption{DCE MRI samples from two (\textbf{(a)i} and \textbf{(b)i}) anonymized patients illustrating distinct sizes and location of breast lesion. The segmentation was performed manually on three slices (top \textbf{a}, middle \textbf{b} and bottom \textbf{c} of breast lesion \textbf{red}). MedSAM was used to create the final mask (\textbf{(a)ii} and \textbf{(b)ii}), and this was implemented into the U-Mamba model, automate the breast lesion regions and effectively capturing these differences between the two cases.}
\label{fig-3}
\end{figure*}

The output of the final U-Mamba segmentation for DCE with tumour lesion identification showed strong performance metrics, as indicated in Table \ref{tab-3}. Mean DSC (+Standard Deviation [SD]) for tumour segmentation was 0.82 (+0.12) MRI samples from two anonymized patients highlighted noticeable variations in the size of the breast lesions (Figure \ref{fig-3}). These differences were clearly reflected in the segmentation results, where the U-Mamba model accurately delineated the differences in breast tumour regions for both cases.  

\textit{\textbf{Merged 3D model}}

\begin{figure*}[t]
\centering
\begin{subfigure}[b]{0.48\textwidth}
    \centering
    \includegraphics[width=\textwidth]{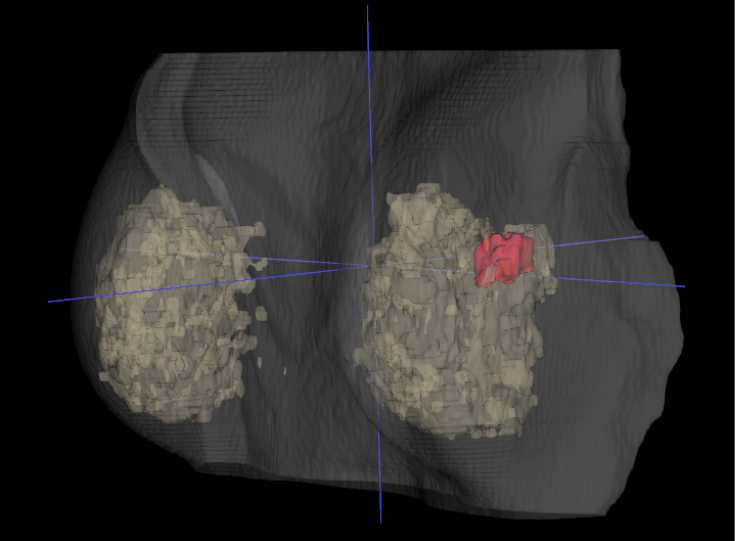}
    \caption{}
    \label{fig:4a}
\end{subfigure}
\hfill
\begin{subfigure}[b]{0.48\textwidth}
    \centering
    \includegraphics[width=\textwidth]{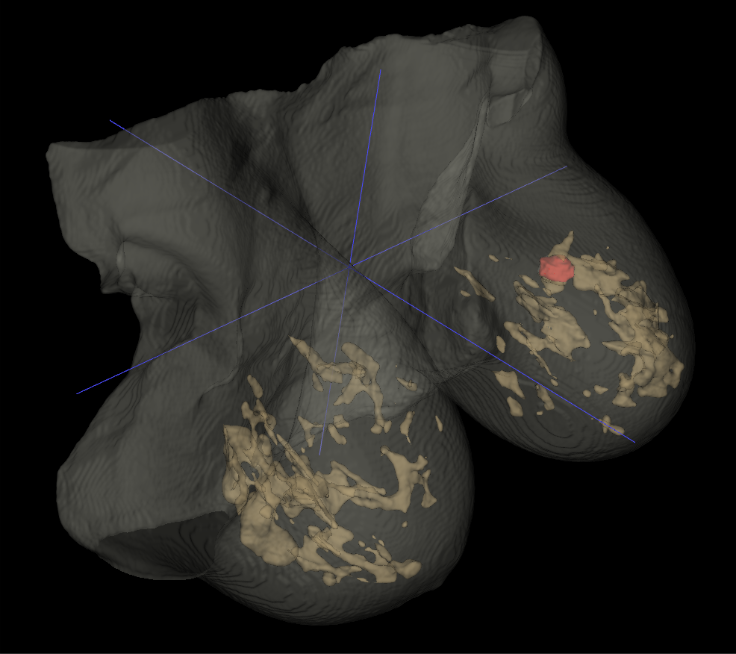}
    \caption{}
    \label{fig:4b}
\end{subfigure}

\caption{3D breast model generated by merging T1W and DCE U-mamba models, illustrating the breast outline, fibroglandular tissue, and lesion. The two examples demonstrate the model's ability to differentiate fibroglandular tissue in relation to the overall breast outline while accurately locating the breast lesion.}
\label{fig-4}
\end{figure*}

The T1W and DCE U-mamba models were merged to create a final 3D representation highlighting the breast outline, fibroglandular tissue and breast lesion, as demonstrated in the two examples (Figure \ref{fig-4}), where the model was able to capture the differences in fibroglandular tissue in relation to the whole breast outline and locate the breast lesion simultaneously. 

\subsection{Qualitative findings}

\textit{Patient semi-structured interviews}

Six patients were interviewed following their exposure to the above models. The qualitative analysis revealed three main themes: 

\begin{enumerate}
    \item Understanding of breast anatomy and disease, where participants expressed that the 3D images significantly improved their comprehension of breast structures, including the distinction between tumour and fibroglandular tissue. For instance, one patient noted, "I finally see what breast cancer looks like", another patient mentioned, "Seeing the tumour in relation to the surrounding tissue made it all feel more real and understandable. 
    \item Clarity on surgical options as most patients reported feeling more informed about the condition, with several expressing a greater awareness of potential surgical interventions. One participant remarked, "Now I can visualize the options such as lumpectomy versus mastectomy much better." 
    \item Empowerment in decision making, where the use of 3D visualisation appeared to enhance patients' confidence in making informed decisions about their treatment. As one patient stated, "I would feel more in control as I can see what I’m dealing with." Another patient said, “the 3D model can help the surgeon explain what they would do during the operation, and this would make to feel more prepared and less scared”. 
\end{enumerate}

\textit{Breast physician semi-structure interviews}

Framework analysis identified three primary themes after interviewing eight breast surgical oncologists: 

\begin{enumerate}
    \item Surgical Planning, where the oncologists noted that the 3D images facilitated a more precise surgical approach, with one surgeon stating, "These images allow us to map the tumour relative to the whole breast”. They emphasized that the detailed visualization of the tumour’s relationship to surrounding tissue improved their ability to plan incisions. 
    \item Intraoperative navigation, where the surgeons reported that 3D reconstruction improves navigation for tumour resection One oncologist remarked, "Having a 3D reference in the OR can help make real-time adjustments, improving both efficiency and safety." 
    \item Patient Education and Shared Decision-Making, highlighting the effectiveness of the images in understanding the concepts of proposed surgery. One participant noted, "These visuals can bridge the gap in understanding, making discussions about treatment options more collaborative." Despite the advantages, the oncologists identified limitations, particularly regarding anatomical accuracy when the images are generated in the prone position, which could lead to misunderstandings about tumour location. 
\end{enumerate}

\section{Discussion} 

The key findings of this study are that we created an ML-based pipeline to semi-autonomously create patient specific 3D breast reconstruction and provide both patient and clinicians visualisations to assist with shared-decision making. To do this we used U-Mamba, a novel architecture for segmentation, to identify the breast outline, fibroglandular tissue, and breast lesion from MRI with strong performance. U-Mamba combines the advantage of convolutional layers and state space models, which can simultaneously capture local features and aggregate long-range dependencies. The study also demonstrates the potential use of the 3D breast model in the clinical setting based on qualitative feedback from both breast surgical oncologists and patients, enabling share-decision making.  

\subsection{Segmentation Models}

The U-Mamba model exhibits robust performance by effectively capturing the varying amounts of fibroglandular tissue present across a wide age range of patients in the study (Table \ref{tab-1}). Its adaptability is demonstrated through consistent accuracy in segmenting both fibroglandular tissue and the whole breast and in various breast T1W images (Figure \ref{fig:workflow}). The model’s ability to differentiate between fibroglandular tissue has been highlighted as an important clinical feature, given the association between dense breast tissue and breast cancer risk.\cite{muller2023fibroglandular, mann2022breast} We also evaluated U-Mamba model on an external dataset, and demonstrated strong performance with our internal dataset by adding human-in-the-loop strategy, compared to evaluation on internal tests only based on previous work \cite{zhang2022automatic, yue2022deep, huo2021segmentation}.  

One significant barrier to the widespread implementation of such models is the lack of standardization in MRI examinations. Different medical centres employ diverse MRI protocols and sequences for diagnosing breast cancer, which complicates the process of developing universal models. This variability in imaging techniques meant that we were unable to identify an existing dataset for dynamic contrast-enhanced (DCE) breast MRI, as the contrast in our images significantly differed from those available in other datasets. Considering that manual segmentation of medical images is highly labor-intensive, we employed MedSAM \cite{ma2024segment} to address this issue. MedSAM offers the ability to accurately identify and delineate key regions in medical images, such as tumours and other tissue abnormalities. The model's architecture is an adaptation of the SAM network. To optimize model performance (Table \ref{tab-2}), we utilized a human-in-the-loop strategy, wherein only the top, middle, and bottom regions of the breast lesion were manually annotated. This was crucial due to the heterogeneity of breast cancer lesions and indistinct boundaries of fibro-glandular tissue \cite{zhuang2019effectiveness}. Recognizing these complexities, we developed bespoke models for both T1W and DCE images, and then combined these models to generate a 3D reconstruction. 

\subsection{Clinical Implications of 3D Models}

Semi-automated 3D segmentation, has been shown  not only assist in tumour annotation, but also to improve the accuracy of T-stage assessment \cite{zhang2023robust}, and estimate breast tumour-to-volume ratio \cite{faermann2014tumor}. 3D technologies also aim to deal with challenging limitations seen in surgery: the lack of realistic visualization methods and the attempt to standardize the patient-specific approach in surgery. 3D reconstruction also helps better prepare surgeons in a risk-free environment \cite{portnoy2023three}, in keeping with our qualitative analysis findings. Guided by 3D reconstruction, surgeons can combine both CT/MR image interpretation and 3D models to develop more advanced skills of mental reconstruction and surgical planning, \cite{yeo2018utility} which was one of the themes identified in our study. Considering the possible healthcare savings, the development of automated segmentation methods and falling technology costs, it would be prudent to quantify the potential benefits of 3D reconstruction in breast surgical practice as the technology becomes easier to adopt. 

3D reconstruction offers substantial value to patient engagement and education \cite{zhuang2019effectiveness, vandebelt2018patient}. We demonstrated that incorporating 3D breast models into patient-shared decision-making plays a crucial role in managing expectations before surgery highlighted by framework analysis. The models can also provide a realistic visualisation, helping clinicians identify patients who may hold unrealistic or misguided expectations about BCS \cite{fuzesi2019expectations}, thus fostering a more accurate understanding of surgical options. It is likely that patient-specific 3D models would be useful tools for securing informed consent for breast surgery. 

\subsection{Limitations}

There are a few limitations to our study. First, we only included mass lesions in this study, and non-mass-like enhancements were excluded, thus our model may not be reliable when encountered with pathologies that do not form mass lesions on MRI, such as ductal-carcinoma in-situ. Second, clinical MR scanning series are diverse, including position reference scan, T2-weighted fat-suppressed scan, T1-weighted non-fat suppressed scan, pre- and post-contrast T1-weighted fat-suppressed scans, raising the generalization problem in image segmentation. Third, we collected MR images scanned in the same hospital introducing bias. In the future, to further evaluate the performance and reliability, we would apply our model on a multicentre dataset. Fourth, we did not segment other regions of the breast, e.g., the nipple areola complex, blood vessels which may assist in surgical planning for optimal cosmetic outcome. Fifth, we did not compare the 3D maximum intensity projection created from the MRIs with our ML-based 3D reconstruction. This could potentially be another method of validating ML-based models in the future. Sixth, we did not investigate inter-rater variability due to the lack of multiple segmentations by multiple readers on the same examinations. Finally, all our 3D models were based on prone-position imaging, as per standard breast MRI protocols, which may alter tumour positioning compared to the supine position.  

\section{Conclusion}

The study demonstrates the development of 3D breast reconstruction models using a novel ML-based approach to autonomously segment fibroglandular tissue and detect lesions from breast MRIs with good performance metrics. Despite the challenge of varying MRI protocols between medical centres based on external dataset, the model performed well through a human-in-the-loop approach. The 3D reconstruction models may offer better visualization for surgeons and help patients understand their surgical options more clearly, aiding in informed decision-making. As 3D reconstruction technology becomes more affordable and easier to use, it could have a significant impact on breast cancer surgery and patient care. 

\newpage
\bibliography{main}

\newpage
\onecolumn
\appendix
\section*{Appendix}
\addcontentsline{toc}{section}{Appendix}

\section{Patient Education Questionnaire}
\label{app:patient-questionnaire}

\textbf{Utility of 3D Breast Reconstruction for Patient Education and Decision-Making}

\subsection{Awareness of 3D Breast Reconstruction}

\begin{enumerate}[label=\alph*.]
\item Were you aware that 3D reconstruction of the breast and tumor anatomy is possible?
\begin{itemize}
\item[$\square$] Yes
\item[$\square$] No
\end{itemize}

\item If yes, how did you learn about it? (Check all that apply)
\begin{itemize}
\item[$\square$] Healthcare provider
\item[$\square$] Media (TV, internet, etc.)
\item[$\square$] Support group
\item[$\square$] Other: \underline{\hspace{3cm}}
\end{itemize}
\end{enumerate}

\subsection{Understanding of 3D Breast Reconstruction}

\begin{enumerate}[label=\alph*.]
\item How well do you understand the concept of 3D breast reconstruction?
\begin{itemize}
\item[$\square$] Very well
\item[$\square$] Somewhat well
\item[$\square$] Not very well
\item[$\square$] Not at all
\end{itemize}

\item Would you like more information about 3D breast reconstruction?
\begin{itemize}
\item[$\square$] Yes
\item[$\square$] No
\end{itemize}
\end{enumerate}

\subsection{Perceived Usefulness of 3D Breast Reconstruction}

\begin{enumerate}[label=\alph*.]
\item Do you believe 3D breast reconstruction can help patients better understand their condition?
\begin{itemize}
\item[$\square$] Yes
\item[$\square$] No
\end{itemize}

\item Do you think 3D breast reconstruction can aid in decision-making regarding treatment options?
\begin{itemize}
\item[$\square$] Yes
\item[$\square$] No
\end{itemize}

\item Would you be interested in viewing a 3D reconstruction of your own breast and tumor anatomy?
\begin{itemize}
\item[$\square$] Yes
\item[$\square$] No
\end{itemize}
\end{enumerate}

\subsection{Preferences for Educational Tools}

\begin{enumerate}[label=\alph*.]
\item Would you prefer traditional 2D images or 3D reconstructions for understanding your condition?
\begin{itemize}
\item[$\square$] Traditional 2D images
\item[$\square$] 3D reconstructions
\end{itemize}

\item What features would you find most helpful in a 3D reconstruction tool? (Check all that apply)
\begin{itemize}
\item[$\square$] Rotating the view of the breast and tumour
\item[$\square$] Zooming in and out for closer examination
\item[$\square$] Annotating areas of interest
\item[$\square$] Other: \underline{\hspace{3cm}}
\end{itemize}
\end{enumerate}

\subsection{Feedback and Suggestions}

\begin{enumerate}[label=\alph*.]
\item Do you have any additional comments, suggestions, or concerns regarding the use of 3D breast reconstruction for patient education and decision-making?

\vspace{3cm}
\end{enumerate}

\noindent Thank you for taking the time to complete this questionnaire. Your feedback is valuable in helping us improve patient care and decision-making processes.

\clearpage

\section{Healthcare Provider Evaluation Form}
\label{app:provider-evaluation}

\textbf{Evaluation Form for 3D Breast Reconstruction Models}

\subsection{Usability}

I found the 3D breast reconstruction model:

\begin{table*}[h]
\centering
\begin{tabular}{p{6cm}ccccc}
\toprule
& \textbf{Strongly Agree} & \textbf{Agree} & \textbf{Neutral} & \textbf{Disagree} & \textbf{Strongly Disagree} \\
\midrule
Is easy to interpret & $\square$ & $\square$ & $\square$ & $\square$ & $\square$ \\
Easy to use in practice & $\square$ & $\square$ & $\square$ & $\square$ & $\square$ \\
Had acceptable resolution & $\square$ & $\square$ & $\square$ & $\square$ & $\square$ \\
Reflects a realistic visual representation of the underlying breast and tumour anatomy & $\square$ & $\square$ & $\square$ & $\square$ & $\square$ \\
\bottomrule
\end{tabular}
\end{table*}

\subsection{Educational Value}

I believe the 3D breast reconstruction model:

\begin{table*}[h]
\centering
\begin{tabular}{p{6cm}ccccc}
\toprule
& \textbf{Strongly Agree} & \textbf{Agree} & \textbf{Neutral} & \textbf{Disagree} & \textbf{Strongly Disagree} \\
\midrule
Is an effective adjunct to current surgical training & $\square$ & $\square$ & $\square$ & $\square$ & $\square$ \\
Is effective at providing pre-operative planning and intra-operative coaching & $\square$ & $\square$ & $\square$ & $\square$ & $\square$ \\
Can improve surgeon performance & $\square$ & $\square$ & $\square$ & $\square$ & $\square$ \\
Can improve patient understanding & $\square$ & $\square$ & $\square$ & $\square$ & $\square$ \\
\bottomrule
\end{tabular}
\end{table*}

\subsection{Adaptation and Integration}

\begin{table*}[h]
\centering
\begin{tabular}{p{6cm}ccccc}
\toprule
& \textbf{Strongly Agree} & \textbf{Agree} & \textbf{Neutral} & \textbf{Disagree} & \textbf{Strongly Disagree} \\
\midrule
Surgical training program incorporation for pre-operative planning and post-operative teaching & $\square$ & $\square$ & $\square$ & $\square$ & $\square$ \\
If available, I would use it consistently for teaching and coaching & $\square$ & $\square$ & $\square$ & $\square$ & $\square$ \\
\bottomrule
\end{tabular}
\end{table*}

\subsection{Design}

\textbf{(Check all that apply)} I prefer the 3D breast reconstruction model:

\subsubsection{Functions}
To offer functions such as:
\begin{itemize}
\item[$\square$] Rotating the view of the breast and tumour
\item[$\square$] Dissecting/cutting tissues
\item[$\square$] Manipulating/retracting tissues
\end{itemize}

\subsubsection{Visualization Options}
To offer different visualization options:
\begin{itemize}
\item[$\square$] Anatomical structures with colour outline
\item[$\square$] Realistic colours from the surgical field
\item[$\square$] Safe and dangerous zones of dissection
\end{itemize}

\subsubsection{Annotation Options}
To offer an annotation option (e.g., freehand drawings):
\begin{itemize}
\item[$\square$] In 2D
\item[$\square$] In 3D
\end{itemize}

\subsubsection{Accessibility}
Be accessible through:
\begin{itemize}
\item[$\square$] A website
\item[$\square$] App that I can download on my personal device
\item[$\square$] Mobile device/Tablet
\item[$\square$] Personal computer
\item[$\square$] Other: \underline{\hspace{3cm}}
\end{itemize}

\subsection{Additional Comments}

Please list any additional comments, suggestions, or considerations:

\vspace{3cm}

\noindent Thank you for your feedback!

\end{document}